\documentclass[10pt, conference, table]{IEEEtran}

\usepackage[]{todonotes}            
\presetkeys{todonotes}{inline}{}  

\usepackage{listings}
\usepackage{xcolor}
\usepackage{siunitx}
\usepackage{multirow}
\usepackage{amsthm}
\usepackage{array,booktabs,arydshln,xcolor}

\usepackage[]{xcolor}

\theoremstyle{definition}

\definecolor{codegreen}{rgb}{0,0.6,0}
\definecolor{codegray}{rgb}{0.5,0.5,0.5}
\definecolor{codepurple}{rgb}{0.58,0,0.82}
\definecolor{backcolour}{rgb}{0.95,0.95,0.92}

\usepackage{mathtools}

\DeclarePairedDelimiter\abs{\lvert}{\rvert}%
\DeclarePairedDelimiter\norm{\lVert}{\rVert}%

\lstdefinestyle{mystyle}{
  backgroundcolor=\color{backcolour},   commentstyle=\color{codegreen},
  keywordstyle=\color{magenta},
  numberstyle=\tiny\color{codegray},
  stringstyle=\color{codepurple},
  basicstyle=\ttfamily\footnotesize,
  breakatwhitespace=false,         
  breaklines=true,                 
  captionpos=b,                    
  keepspaces=true,                 
  numbers=left,                    
  numbersep=5pt,                  
  showspaces=false,                
  showstringspaces=false,
  showtabs=false,                  
  tabsize=2
}

\usepackage{cite}
\usepackage{hyperref}

\ifCLASSINFOpdf
\else
\fi

\usepackage[boxruled]{algorithm2e}

\usepackage{url}

\begin{document}
\title{On Metrics to Assess the Transferability of Machine Learning Models in Non-Intrusive Load Monitoring}

\author{\IEEEauthorblockN{Christoph Klemenjak}
\IEEEauthorblockA{University of Klagenfurt\\
Klagenfurt, Austria\\
klemenjak@ieee.org}
\and
\IEEEauthorblockN{Anthony Faustine}
\IEEEauthorblockA{University of Ghent\\
Ghent, Belgium\\
sambaiga@gmail.com}
\and
\IEEEauthorblockN{Stephen Makonin}
\IEEEauthorblockA{Simon Fraser University\\
Burnaby, Canada\\
smakonin@sfu.ca}
\and
\IEEEauthorblockN{Wilfried Elmenreich}
\IEEEauthorblockA{University of Klagenfurt\\
Klagenfurt, Austria\\
wilfried.elmenreich@aau.at}}

\maketitle

\begin{abstract}
To assess the performance of load disaggregation algorithms it is common practise to train a candidate algorithm on data from one or multiple households and subsequently apply cross-validation by evaluating the classification and energy estimation performance on unseen portions of the dataset derived from the same households.
With an emerging discussion of transferability in Non-Intrusive Load Monitoring (NILM), there is a need for domain-specific metrics to assess the performance of NILM algorithms on new test scenarios being unseen buildings.
In this paper, we discuss several metrics to assess the generalisation ability of NILM algorithms. These metrics target different aspects of performance evaluation in NILM and are meant to complement the traditional performance evaluation approach. We demonstrate how our metrics can be utilised to evaluate NILM algorithms by means of two case studies. We conduct our studies on several energy consumption datasets and take into consideration five state-of-the-art as well as four baseline NILM solutions. Finally, we formulate research challenges for future work.
\end{abstract}
\begin{IEEEkeywords}
NILM, Load Disaggregation, Energy Disaggregation, Generalisation, Transferability, Performance Evaluation
\end{IEEEkeywords}

\IEEEpeerreviewmaketitle



\section{Introduction}

The worldwide adoption of smart meters is expected to produce significant quantities of energy consumption data at an unprecedented rate. The analysis of these data streams can play a vital role in the design of customised energy
efficiency and energy demand management strategies, improving electricity system optimisation and enhance system monitoring~\cite{Cominola2017}.

Non-Intrusive Load Monitoring (NILM), also known as load disaggregation, is a technique for analysing data extracted from single point sources such as smart meters. It relies on signal processing and machine learning algorithms to infer ON/OFF states and estimate power consumption of one or more appliances from only the aggregated power data. Therefore, only one meter, installed at a single point, is required, while the household appliances to be monitored do not need to be equipped with metering devices \cite{hart1992nonintrusive,klemenjak2018yomopie,harell2019wavenilm}. Compared to other monitoring technologies, NILM can easily be integrated in existing buildings without causing inconvenience to inhabitants due to the installation process~\cite{faustine2017survey}.

Mathematically, we describe NILM as the problem of providing estimates $[\hat{x}_t^{(1)}, \dots ,\hat{x}_t^{(M)}]$ of the actual power consumption of $M$ electrical appliances $[x_t^{(1)}, \dots ,x_t^{(M)}]$ at time $t$ given only the aggregated power consumption series $y_t$. We refer to algorithms applied to solve load disaggregation problems as load disaggregation algorithms. The aggregate power signal $y_t$, provided to a load disaggregation algorithm consists of
\begin{equation}
	y_t = \epsilon_t + \sum_{i=1}^{M}{x_t^{(i)}}
\end{equation}
with $M$ appliance-level signals $x_t^{(i)}$ and an error term $\epsilon_t$, which models the discrepancy between the sum of the individual measurements and the overall branch measurement \cite{klemenjak2020comparability}. 

Machine learning approaches such as Deep Learning techniques~\cite{kelly2015neural,zhang2018sequence} and Hidden Markov Model (HMM)~\cite{kolter2010energy,makonin2015nonintrusive} have achieved state-of-the-art performance in NILM. Even though state-of the-art NILM algorithms offer better performance, they either require vast amounts of labelled data or additional prior information for training. This drawback limits the applicability of NILM in many real-world situations, as careful annotation and labelling of data is costly and time-consuming. Another challenge is to generalise NILM  models across datasets or buildings~\cite{Liu2019}.

To address the above challenges, it is important to build NILM models with strong transferability. Meaning, the ability of a model developed for a specific dataset or house to be generalised on new houses not present in the training set. Recently, it has been demonstrated that deep learning~\cite{murray2019transferability} and model-based NILM~\cite{Humala2018} trained on a specific dataset can be transferred to other datasets/buildings with minimal performance drop. However, it is not clear how to quantify transferability of machine learning models in NILM. There is a strong need for domain-specific metrics as well as a clear evaluation methodology.


Relying on a single dataset during the development of a NILM algorithm results in issues like overfitting and a lack of generalisation \cite{klemenjak2019electricity}. With an upcoming discussion of transferability in NILM, we observe that most studies test transferability on only a few unseen houses ~\cite{Gao2018,kelly2015neural,murray2019transferability}. This may be the result of a lack of domain-specific metrics to evaluate transferability as well as a clear methodology to evaluate generalisation abilities of NILM algorithms.

In this paper, we aim to close this gap by introducing domain-specific transferability metrics. We discuss what strategies exist to evaluate generalisation abilities of NILM algorithms. With paramount importance being the consideration of how many unseen test scenarios (buildings) are used as part of the evaluation of transferability. Based on this taxonomy, we introduce four metrics that cover different aspects of the transferability problem. To demonstrate how the metrics are meant to be applied in future investigations, we present two case studies that examine generalisation abilities of four baseline and five advanced NILM algorithms. Finally, we conclude the paper by formulating research challenges for future work.

It should be noted that while the focus of this discussion is on NILM, we expect that the presented metrics can be applied to related Machine Learning problems such as forecasting of energy consumption or classification problems in smart grids. To foster reproducibility of experiments and to allow future investigators to make use of our implementation, we provide main parts of the source code used in our investigations on a GitHub repository\footnote{\url{https://github.com/klemenjak/nilm-transferability-metrics}}.


\section{Evaluation Strategies} \label{sec:strategies}

To date, the majority of studies in NILM scholarship has mainly focused on testing on seen buildings where an algorithm is tested on data of a building that the algorithm has seen before during training. In this way, researchers evaluate how well the trained algorithm is able to detect one particular appliance instead of an appliance type. 
With regard to transferability of appliance models, we claim that appliance models should be tested on one or multiple unseen households. In this way, it can be assessed how well an algorithm generalises to appliances of the same kind. We identify several evaluation strategies that differ in the number of seen and unseen houses:
\begin{itemize}
    \item \emph{1-to-1 testing}: An algorithm is trained on household A. During evaluations, one test on household A (a seen test) and one test on a hitherto unseen household B (an unseen test) is conducted. As a result, we obtain the performance on one seen and one unseen household, which provides a simple estimate on the generalisation ability of the algorithm.
    
    \item \emph{1-to-N testing}: An algorithm is trained on household A. The performance of the algorithm is assessed on the seen household A and N hitherto unseen households, where $\text{N} > 1$. In this case, a broader understanding of the generalisation ability of the algorithm on new unseen test cases is obtained.

    \item \emph{M-to-N testing}: An algorithm is trained on data from M households. In the course of experiments, the performance of the algorithm on the M households seen during training and N other unseen households is assessed. 
    
\end{itemize}
In order to report how comprehensively generalisation abilities were evaluated in an experiment, we identify the need for a domain-specific metric taking into account the number of seen and unseen households and being applicable to all types of NILM experiments. 


With the \emph{Generalisation Ratio (GR)}, we introduce a metric to do serve this purpose. The GR is defined as the ratio between the number of seen households and the number of unseen households during an experiment:

\begin{equation}
    \text{GR} = \text{\#seen tests : \#unseen tests}
\end{equation}
This metric is defined under the assumption that experiments on households are independent observations. 

Table \ref{tab:genratio} demonstrates how the generalisation ratio is found for a number of sample experiments. The table lists six experiments with different numbers of seen and unseen tests. As the examples show, the generalisation ratio summarises in a clear manner how many seen and unseen tests were performed in an experiment. Furthermore, the GR instantly points out what type of evaluation strategy has been applied (1-to-1, 1-to-N, M-to-N, etc.).

\begin{table}[!t]
\renewcommand{\arraystretch}{1.3}
\caption{Generalisation Ratio of selected NILM experiments}
\label{tab:genratio}
\centering
\begin{tabular}{|c|c|l|c|}
\hline
 Training      & Seen Tests    & Unseen Tests                                                                       & GR \\ \hline
 REFIT house 1 & REFIT house 1 & None                                                                                         & 1:0  \\ 
\rowcolor[HTML]{EFEFEF} 
 DRED house 1  & DRED house 1  & ECO house 2                                                                                  & 1:1  \\ 
 \begin{tabular}[c]{@{}l@{}}ECO house 1, 4\\ REDD house 1\end{tabular}   & \begin{tabular}[c]{@{}l@{}}ECO house 1, 4\\ REDD house 1\end{tabular}   & \begin{tabular}[c]{@{}l@{}}ECO house 2\\ REDD house 6\end{tabular}                           & 3:2  \\ 
\rowcolor[HTML]{EFEFEF} 
REDD house 1  & REDD house 1  & \begin{tabular}[c]{@{}l@{}}REDD house 2, 3\\ REFIT house 4, 5, 9\\ SynD house 1\end{tabular} & 1:6  \\
 SynD house 1  & SynD house 1  & \begin{tabular}[c]{@{}l@{}}REDD house 1, 2, 3\\ REFIT house 2, 6, 9, 15\\ ECO house 5\end{tabular} & 1:8  \\ 
\rowcolor[HTML]{EFEFEF} 
 SynD house 1  & SynD house 1  & \begin{tabular}[c]{@{}l@{}}REDD house 1, 3, 4\\ REFIT house 2, 3, 10, 11\\ ECO house 1\end{tabular} & 1:8  \\ 
\hline
\end{tabular}
\end{table}

\section{Generalisation Loss} \label{sec:genloss}

Performance evaluation in NILM focuses on how accurately a NILM algorithm can estimate each appliance state and how much power it consumes \cite{klemenjak2020comparability}. For this reason, we claim that transferability metrics have to take into account accuracy on unseen houses. 
 
In NILM, we observe two main approaches to solve the disaggregation problem: event detection (ED) and energy estimation (EE) \cite{pereira2017comparison}.
Therefore, it is evident that transferability metrics have to be defined in a broad sense and preferably complement the traditional evaluation approach rather than representing an alternative approach. 

Let us consider event detection approaches first. We advocate that a domain-specific metric for transferability should link accuracy on seen houses and accuracy on unseen houses. Motivated by this, we introduce the metric \emph{generalisation loss (G-loss)}, which is defined as
\begin{equation}
     \text{G-loss} = 100 \cdot (1 - \frac{\text{ACC}_u}{\text{ACC}_s}) \text{ \%}
\end{equation}
the loss in accuracy with regard to the accuracy on a seen house $\text{ACC}_s$ and the accuracy on an unseen house $\text{ACC}_u$. The generalisation loss is defined for accuracy metrics that take on values in the interval $[0, 1]$ (e.g., Accuracy, F1-Score, Hamming Loss, etc.) and only for experiments where $\text{ACC}_s > 0$. For such accuracy metrics, the G-loss takes on values between 0 and 100 \%. A G-loss of 70\% reports that, with regard to an accuracy metric $\text{ACC}$, the accuracy on the unseen house is 70\% lower than the accuracy on the seen house. Therefore, the G-loss represents a simple measure to evaluate generalisation abilities for 1-to-1 transferability tests. 

The second group of NILM approaches comprises energy estimation approaches. These approaches provide estimates of the actual power consumption of appliances. To assess the accuracy of such approaches, common regression metrics are used for the large part. In contrast to classification metrics, the majority of regression metrics reports either relative or absolute errors. The difference lies in the best possible value, which is 0 for the most common regression metrics in NILM such as RMSE, MAE, NEP, and NDE \cite{pereira2018performance}. For these common regression metrics, we define the \emph{generalisation loss (G-loss)} as
\begin{equation}
     \text{G-loss} = 100 \cdot (\frac{\text{ERR}_u}{\text{ERR}_s} - 1)  \text{ \%}
\end{equation}
the increase in error where ${\text{ERR}_s}$ is the error on the seen house and ${\text{ERR}_u}$ the error on the unseen house. The G-loss is defined for experiments where $\text{ERR}_s >0$. With regard to a regression metric ${\text{ERR}}$, a G-loss of 50\% indicates that the error on the unseen house is 50 \% larger than the error on the seen house.

For both, event detection and energy estimation approaches, the generalisation loss serves to complement the traditional way of performance evaluation as it is related to one specific metric a time. We illustrate the intended use of this metric in Table \ref{tab:gen-loss-exps}. We present the outcome of three independent NILM experiments. In every experiment, we trained and tested a NILM algorithm on house 1 of a dataset. An accuracy $\text{ACC}_s$ of 0.98 was observed in experiment 1, an F1-Score $\text{ACC}_s$ of 0.91 in experiment 2, and a MAE error $\text{MAE}_s$ of \SI{30.81}{\watt} in experiment 3. Those results represent the performance of the algorithm on the seen household (result of the seen test). In all three experiments, the trained algorithm was tested on houses 2 to 9 with $\text{ACC}_u$ in experiment 1, $\text{F1}_u$ in experiment 2, and $\text{MAE}_u$ in experiment 3. We observe two things in particular: First, we see how the generalisation loss points out how the performance on the unseen household relates to the performance on the seen household. Second, we see that the G-loss is associated with one metric per study: The loss of the accuracy in experiment 1, the loss of the F1-score in experiment 2, and the increase of the MAE in experiment 3.
\begin{table}[]
\renewcommand{\arraystretch}{1.3}
\caption{Generalisation loss for common NILM metrics}
\label{tab:gen-loss-exps}
\centering
\begin{tabular}{|c|cc|cc|cc|}
\hline
  & \multicolumn{2}{c|}{Experiment 1}    & \multicolumn{2}{c|}{Experiment 2} & \multicolumn{2}{c|}{Experiment 3} \\
  & \multicolumn{2}{c|}{$\text{Acc}_s$= 0.98}    & \multicolumn{2}{c|}{$\text{F1}_s$= 0.91 } & \multicolumn{2}{c|}{$\text{MAE}_s$= \SI{30.81}{\watt}} \\
House & $\text{Acc}_u$         & G-loss  & $\text{F1}_u$         & G-loss      & $\text{MAE}_u$   & G-loss          \\ \hline
2      & 0.74 & 24.5 \% & 0.69                &  24.2   \%           &   \SI{54.78}{\watt}  &    77.8 \%              \\
\rowcolor[HTML]{EFEFEF} 
3      & 0.60 & 38.8 \% &  0.55              & 39.6  \%             &   \SI{39.62}{\watt} &  28.6  \%             \\
4      & 0.77 & 21.4 \% &    0.72              & 20.9   \%            &   \SI{33.12}{\watt}  &     7.5 \%             \\
\rowcolor[HTML]{EFEFEF} 
5      & 0.37 & 62.2 \% &   0.32              & 64.8  \%             &   \SI{45.39}{\watt}  &     47.3  \%           \\
6      & 0.86 & 12.2 \%  &    0.81              & 11.0   \%            &   \SI{50.59}{\watt}   &    64.2 \%             \\
\rowcolor[HTML]{EFEFEF} 
7      & 0.17 & 82.7  \% &    0.12              & 86.8  \%             &   \SI{38.73}{\watt}   &    25.7  \%            \\
8      & 0.06 & 93.9  \% &    0.01              & 98.9  \%             &   \SI{60.54}{\watt}     &  96.5 \%               \\
\rowcolor[HTML]{EFEFEF} 
9      & 0.88 & 10.2  \% &    0.83              &     8.8 \%           &   \SI{41.14}{\watt}    &    33.5   \%   \\   \hline    
\end{tabular}
\end{table}

\begin{table}[]
\renewcommand{\arraystretch}{1.3}
\caption{ Assessing Transferability in Experiments }
\label{tab:gen-loss}
\centering
\begin{tabular}{|c|lccc|}
\hline
Experiment & Seen House      & Unseen Houses & MGL & GR \\ \hline
1          & Accuracy = 0.98 & AUH = 0.56    &  42.86 \%   &  1:8  \\
\rowcolor[HTML]{EFEFEF} 
2          & F1 = 0.91       & AUH = 0.51    &  43.96 \%   &  1:8  \\
3          & MAE = \SI{30.31}{\watt}    &  EUH = \SI{45.49}{\watt}    & 47.65 \%    &  1:8  \\ \hline
\end{tabular}
\end{table}

\section{Performance on Unseen Buildings} \label{sec:unseen}

Evaluating the performance of a NILM algorithm on several unseen households provides a better estimate of its generalisation abilities, as the experiments in Table \ref{tab:gen-loss-exps} indicate. For such 1-to-N experiments, we introduce the metric \emph{mean generalisation loss (MGL)}. This metric takes into account the generalisation loss at every unseen house:
\begin{equation}
    \text{MGL} = \frac{1}{N} \sum_{i=1}^{N}{\text{G-loss}_i}
\end{equation}
where $\text{G-loss}_i$ represents the generalisation loss at building i and N is the number of unseen houses during the experiment. This metric reports the mean generalisation loss on unseen houses in a 1-to-N NILM experiment. 

Besides performance loss, we maintain that the actual performance of NILM algorithms on unseen houses is of interest. To quantify the performance of an algorithm on unseen houses (i.e., to measure how well the algorithm generalises) we define the  \emph{accuracy on unseen houses (AUH)} for classification metrics and the \emph{error on unseen houses (EUH)} for regression metrics. Both, AUH and EUH, are found by
\begin{equation}
    \text{AUH} = \frac{1}{N} \sum_{i=1}^{N}{\text{ACC}_{u_i}} \text{ , }\text{EUH} = \frac{1}{N} \sum_{i=1}^{N}{\text{ERR}_{u_i}}
\end{equation}
averaging the performance on individual unseen houses $\text{ACC}_{u_i}$ and $\text{ERR}_{u_i}$. In contrast to MGL, these metrics can be computed for any NILM experiment (1-to-N as well as M-to-N) testing.

Based on the results of Table \ref{tab:gen-loss-exps}, we summarise how AUH, EUH, MGL and GR can be used to examine the generalisation abilities of NILM algorithms in Table \ref{tab:gen-loss}.

The table shows how our approach complements the traditional performance evaluation approach by summarising performance on unseen houses. While we observe excellent values for accuracy, F1-score and MAE on seen houses, the metrics AUH, EUH and MGL reveal that the algorithms show poor generalisation abilities in all three experiments. For instance: In experiment 1, we observe an F1-score of 0.98 on the seen house. AUH signals that the average F1-score of this algorithm on unseen households in experiment 1 is just 0.56, which can be interpreted as poor classification performance. Furthermore, MGL records a mean loss in accuracy of 42.86 \% between seen and unseen tests. The conclusion is that while the algorithm in experiment 1 shows excellent classification performance on the seen household, the performance on the eight unseen households can be categorised as poor. We make similar observations in experiment 2 and 3. The cause of this poor transferability could be for instance overfitting. It should be noted that the generalisation ratio (GR) is 1:8 for all experiments in this study; i.e., 1-seen test vs. 8-unseen tests. 


\section{Case Studies} \label{sec:experiments}
\begin{table*}
 \begin{minipage}{0.5\linewidth}
 \centering
\renewcommand{\arraystretch}{1.3}
\caption{Case study on generalisation of fridge models}
\label{tab:cross-fridge}
\centering
\begin{tabular}{|l|ccc|ccc|}
\hline
          & \multicolumn{3}{c|}{Classification Accuracy} & \multicolumn{3}{c|}{Estimation Accuracy}  \\
Algorithm & $\text{F1}_s$     & AUH    & MGL   & $\text{MAE}_s$          & EUH         & MGL      \\ 
            & - & - & {[}\%{]}&{[}W{]} &{[}W{]} & {[}\%{]}  \\\hline
CO         &   0.82    &  0.55      &  32.76    &     8.86        &   51.61   & 482.53          \\
\rowcolor[HTML]{EFEFEF} 
DAE         &  0.93     &  0.51      &  45.76    &   3.11          &   43.12          & 1287.32         \\
DSC      &  0.33     &  0.46      &  -37.24     &    39.94        &  78.77          &    97.24       \\
\rowcolor[HTML]{EFEFEF} 
FHMM      &  0.83     &  0.55      &  33.49     &  8.52           &  51.38           & 503.52           \\
Hart85      &  0.31     & 0.56       &  -78.74     &   40.29          &   45.45          &  12.81          \\
\rowcolor[HTML]{EFEFEF} 
RNN      &  0.86     &   0.58     & 33.14      &    5.85         &    46.18         &  689.93         \\
Seq2Point      &    0.97   &  0.42      &  57.22     &  1.63           &  42.98           & 2533.98           \\
\rowcolor[HTML]{EFEFEF} 
Seq2Seq          &  0.96     &  0.49      & 48.46     &   2.18          &   42.90          & 1866.66          \\ 
W-GRU          &  0.98     &  0.61      &  38.10    &   1.65          &    40.91         & 2367.39          \\ \hline
\end{tabular}

 \end{minipage}%
 \begin{minipage}{0.5\linewidth}
  \centering
\renewcommand{\arraystretch}{1.3}
\caption{Case study on generalisation of washing machine models}
\label{tab:cross-washer}
\centering
\begin{tabular}{|l|ccc|ccc|}
\hline
          & \multicolumn{3}{c|}{Classification Accuracy} & \multicolumn{3}{c|}{Estimation Accuracy}  \\
Algorithm & $\text{F1}_s$    & AUH    & MGL   & $\text{MAE}_s$         & EUH         & MGL       \\ 
            & - & - & {[}\%{]}&{[}W{]} &{[}W{]} & {[}\%{]}  \\ \hline
CO         &   0.16    &   0.18     &  -9.53    &   232.8          &   176.8          & -24.04        \\
\rowcolor[HTML]{EFEFEF} 
DAE         &  0.37     &   0.16     &  56.83    &    12.89         &   80.94          & 527.58          \\
DSC      &  0.26    &  0.19     & 27.28   &   52.91         &  104.4          &    97.25        \\
\rowcolor[HTML]{EFEFEF} 
FHMM      &  0.32     &  0.17      & 46.78      &   62.99          &   244.1          & 287.52            \\
Hart85      &  0.15     &  0.18      &  -17.94     &    49.84         &   67.78          & 35.99         \\
\rowcolor[HTML]{EFEFEF} 
RNN      &  0.83     &  0.11      &  86.55     &   2.17          &   78.96          &  3534.61         \\
Seq2Point      &  0.89     & 0.22       &   74.74    &  4.15           &    56.50         & 1259.37          \\
\rowcolor[HTML]{EFEFEF} 
Seq2Seq          &   0.46    &  0.14      & 68.47     &   7.02          &   62.92          & 796.65         \\ 
W-GRU          &  0.38     &  0.10      &  71.18    &   11.79          &  109.9           & 832.55         \\ \hline
\end{tabular}
 \end{minipage}%
\end{table*}


We present two case studies to demonstrate how our metrics can be used to assess generalisation abilities of nine state-of-the-art NILM algorithms. 
To allow other investigators to reproduce our results, we utilise the latest version of NILMTK \cite{batra2014nilmtk} and the majority of algorithms provided by NILMTK-contrib \cite{batra2019towards}. Our studies include four simple benchmark algorithms: combinatorial opitmisation (CO) \cite{batra2015if} , discriminative sparse coding (DSC) \cite{kolter2010energy}, exact factorial hidden markov model (FHMM) \cite{batra2014nilmtk}, an implementation of Hart's algorithm (Hart85) \cite{batra2014nilmtk}, \cite{hart1992nonintrusive} and five advanced NILM algorithms: denoising autoencoder (DAE) \cite{kelly2015neural}, recurrent neural network (RNN) \cite{kelly2015neural}, Sequence-to-Point \cite{zhang2018sequence}, Sequence-to-Sequence \cite{zhang2018sequence}, online gated recurrent units (W-GRU) \cite{krystalakos2018sliding}.
We incorporate households from the datasets ECO \cite{beckel2014eco}, REDD \cite{kolter2011redd}, REFIT \cite{murray2017electrical}, and SynD\footnote{\url{https://github.com/klemenjak/SynD}} to train and test our algorithms.
As suggested in \cite{makonin2015nonintrusive}, we take into consideration both: estimation and classification accuracy.
To measure classification performance of algorithms, we apply the F1-score (F-measure):
\begin{equation}
    \text{F1} = 2 \cdot \frac{ \text{precision}  \cdot \text{recall} }{\text{precision}  + \text{recall}} 
\end{equation}
To quantify the disaggregation error, we utilise the mean-absolute error (MAE) between the ground-truth signal $x_i$ and the estimated power consumption $x_i$ of appliance $i$.
\makeatletter
\let\oldabs\abs
\def\abs{\@ifstar{\oldabs}{\oldabs*}}
\let\oldnorm\norm
\def\norm{\@ifstar{\oldnorm}{\oldnorm*}}
\makeatother
\newcommand*{\Value}{\frac{1}{2}x^2}%
\begin{equation}
    \text{MAE} = \frac{1}{T} \cdot \sum_{t=0}^{T-1}{\abs*{ \hat{x}_t^{(i)}-x_t^{(i)}}}  
\end{equation}

We apply the same methodology for both case studies:  We divide data from house 1 of SynD into a training and test set, train our algorithms on the training set and evaluate the performance of our algorithms on the test set. The outcome of this step is the classification accuracy $\text{F1}_s$ and the estimation accuracy $\text{MAE}_s$ of the seen household. Next, we evaluate the classification $\text{F1}_{u_i}$ as well as estimation accuracy $\text{MAE}_{u_i}$ on data from eight unseen households from other datasets.  Based on these accuracy\rq{}s, we compute the accuracy on unseen households (AUH and EUH) and the mean generalisation loss (MGL) for every of our nine algorithms. It should be noted that the test sets of the unseen households have identical properties as the test set in SynD: we tested for a duration of 14 days and applied a sampling interval of \SI{10}{\second} in all training as well as test sets.

Table \ref{tab:cross-fridge} summarises the outcome of our first case study. In this study, we used data of the fridge in house 1 of SynD to train our algorithms. The second last row of Table \ref{tab:genratio} summarises which households from other datasets served as unseen households. For the majority of algorithms in this study, we observe a high classification accuracy $\text{F1}_s > 0.80$ and a low energy estimation errors $\text{MAE}_s < \SI{6}{\watt}$ on the seen household. As concerns the accuracy on unseen houses, we observe that AUH is considerably lower than $\text{F1}_s$ for all algorithms; i.e., the classification accuracy of our algorithms is much lower on unseen houses than on the seen house. 
We make similar observations with regard to the estimation accuracy: for every single algorithm in our study, the error on unseen houses (EUH) is multiple times higher than the mean absolute error on the seen house $\text{MAE}_s$. The mean generalisation loss (MGL) clearly approves this observation, as we observe losses up to 2533.98 \%. 
Contrary to other algorithms, DSC and Hart85 show better classification performance on unseen houses than on the seen house. We also observe how the generalisation loss points this out by being negative, which would correspond to performance gain. However, it has to be pointed out that neither the accuracy on the seen nor the unseen households is satisfactory, as all scores rank clearly below 0.6. 

We summarise the outcome of our second case study in Table \ref{tab:cross-washer}. In this study, we selected data from the washing machine in house 1 of SynD to serve as appliance of interest. The last row of Table \ref{tab:genratio} summarises which households from other datasets served as unseen households. As concerns classification accuracy on unseen households, we observe even lower classification accuracy than in the previous study. Also on the seen household, we observe poor F1-scores for seven of nine algorithms. We observe a similar situation concerning the estimation accuracy. While the estimation performance on the seen household is acceptable for advanced NILM algorithms, the error on unseen households (EUH) is considerably higher, as the MGL confirms with increases of up to 3234.61 \%. This study shows how our metrics can be used to identify overfitted algorithms. For instance: The RNN in this study shows good classification as well as energy estimation performance on the seen household. As concerns traditional performance assessment, the RNN would be classified as good NILM approach. In contrast, when taking the transferability aspect into account, it is evident that the network suffers from overfitting to the washing machine in SynD, as AUH, EUH and MGL clearly point out in our study. 

Based on observations made in our case studies, we draw several conclusions for future work: First, the metrics AUH, EUH, and MGL complement the traditional way of evaluating performance in NILM, as they base on traditional classification and regression metrics. However, these metrics \emph{must always} be reported in conjunction with the performance metric they refer to. Second, generalisation ratio (GR), accuracy on unseen houses (AUH) and error on unseen houses (EUH) can be computed for any NILM experiment. Third, our metrics can be used to identify overfitting.

\section{Related Work} \label{sec:related}

Ensuring transferability of machine learning algorithms is vital for successful adoption and deployment of NILM~\cite{Liu2019}. However, quantifying the generalisation abilities of NILM algorithms is an area that has received little attention in the NILM research community. Training machine learning models on a specific set of buildings and testing them on a new set of buildings not seen during training has become the standard way of evaluating the transferability of NILM models~\cite{zhang2018sequence,kelly2015neural,Gao2018}. Recently, it has been demonstrated that NILM models can be trained on one dataset and be transferred to another dataset with minimal performance drop, if certain requirements are fulfilled~\cite{Humala2018,murray2019transferability}.

Two transfer learning schemes for NILM, appliance transfer learning (ATL) and cross-domain transfer learning (CTL), are proposed in \cite{dincecco2019transfer}. The authors investigate in transferability of Deep Neural Networks for NILM and find that sequence-to-point learning is transferable in the sense that this technique can be applied to test data without fine-tuning, given that the training and test data are in a similar domain. Unfortunately, the authors neither elaborate problem-specific metrics nor perform tests on several unseen houses per experiment. 

Studies on neural network architectures for cross-dataset transferability are presented in \cite{murray2019transferability}. The authors present two architectures, one based on CNN and the other on GRUs, and demonstrate transferability between single households. As the emphasis of the work is on 1-to-1 tests, only conventional accuracy metrics are used in their studies. Nevertheless, the architectures tested in those studies show improved generalisation abilities when compared to the state of the art.

In \cite{murray2019transferability,Humala2018}, transferability of deep learning models is quantified by general disaggregation metrics such as $F_1$ score, accuracy and MSE loss. The problem with such conventional metrics is that they do not take into account the number of seen and unseen houses. We argue that generalisation metrics 
should either link the performance on seen and unseen scenarios or give performance estimates for unseen environments. To address the above challenge, a measure for quantifying transferability of NILM algorithms is proposed in~\cite{nalmpantis2019machine}. Basically, this metric is defined as the standard deviation of the total disaggregation accuracy for various houses. However, the proposed metric only measures how much the performance varies but leaves out from what centre the spread is measured. Thus, we claim that finding metrics that quantify the generalisation ability of NILM models is still an open problem.

\section{Conclusion} \label{sec:conclusion}

In this paper, we introduce several metrics for the assessment of generalisation abilities of NILM algorithms: generalisation ratio (GR), Accuracy on Unseen Houses (AUH), Error on Unseen Houses (EUH), and Mean Generalisation Loss (MGL). The introduced metrics are meant to complement the conventional performance evaluation approach, as they take into consideration the performance of NILM algorithms on unseen houses. Besides performance on new test scenarios, researchers can utilise those metrics to identify overfitted appliance models. We present two case studies to demonstrate how our metrics can be applied to assess generalisation abilities. We examine classification as well as energy estimation accuracy for five state of the art and four baseline NILM algorithms on several datasets. Inspired by the outcome of our studies, we advise future investigators to incorporate at least generalisation ratio (GR) and accuracy on unseen houses (AUH/EUH) into their evaluations. Finally, we identify challenges for future work:

\begin{itemize}
    \item While the metrics GR, AUH and EUH can be used in any evaluation scenario, the mean generalisation loss is only defined for 1-to-1 and 1-to-N testing scenarios -- MGL takes into account only one seen household. Future work should investigate how this metric could be adapted for M-to-N testing scenarios, as most researchers train algorithms on multiple houses.
    \item Examine transferability of new NILM solutions to real use cases such as \cite{jones2020increasing}
    \item Investigate how other accuracy metrics such as the Matthews Correlation Coefficient (MCC) can be applied to evaluate generalisation
    \item Perform extensive case studies on the generalisation abilities of state of the art NILM algorithms
    \item Explore how existing NILM algorithms could be improved with regard to generalisation. 
\end{itemize}


\appendix

The NILM toolkit (NILMTK) allows reproducible experiments and integrates several state-of-the-art algorithms. This allows to repeat studies of related work in a convenient and fast manner. We provide an implementation of our transferability metrics for the latest major release of NILMTK. Our code makes use of NILMTK's API as it analyses the outcome of NILMTK experiments and returns a Python dictionary that summarises the generalisation abilities of tested algorithms.

\lstset{style=mystyle}
\begin{lstlisting}[language=Python, label=lst:usage]
from GenLoss import *
from nilmtk.api import API
from file_handler import load_experiment

from nilmtk.disaggregate import FHMMExact, Hart85
from nilmtk_contrib import *

# load dict and execute experiment
experiment = load_experiment(experiment_ID)
api_results = API(experiment)

# assess generalisation abilities
g_loss = mean_generalization_loss(api_results)
auh = accuracy_on_unseen_houses(api_results)

exit()
\end{lstlisting}

\end{document}